\documentclass[runningheads]{llncs}
\usepackage{graphicx}
\usepackage{comment}
\usepackage{booktabs}
\usepackage{amsmath,amssymb} % define this before the line numbering.
\usepackage{color}

\usepackage[width=122mm,left=12mm,paperwidth=146mm,height=193mm,top=12mm,paperheight=217mm]{geometry}

\begin{document}
% \linenumbers
\pagestyle{headings}
\mainmatter
\def\ECCVSubNumber{2772}  % Insert your submission number here

\title{Sub-Image Anomaly Detection with Deep Pyramid Correspondences} % Replace with your title


\titlerunning{Sub-Image Anomaly Detection with Deep Pyramid Correspondences}
\author{Niv Cohen \and Yedid Hoshen}
\authorrunning{Niv Cohen and Yedid Hoshen}
\institute{
School of Computer Science and Engineering \\ The Hebrew University of Jerusalem, Israel.\\
\email{\{niv.cohen2,yedid.hoshen\}@mail.huji.ac.il}}

%******************
\maketitle

\begin{abstract}
Nearest neighbor (kNN) methods utilizing deep pre-trained features exhibit very strong anomaly detection performance when applied to entire images. A limitation of kNN methods is the lack of segmentation map describing where the anomaly lies inside the image. In this work we present a novel anomaly segmentation approach based on alignment between an anomalous image and a constant number of the similar normal images. Our method, Semantic Pyramid Anomaly Detection (SPADE) uses correspondences based on a multi-resolution feature pyramid. SPADE is shown to achieve state-of-the-art performance on unsupervised anomaly detection and localization while requiring virtually no training time.
\end{abstract}

\section{Introduction}
\label{sec:intro}

Humans observe many images throughout their lifetimes, most of which are of little interest. Occasionally, an image indicating an opportunity or danger appears. A key human ability is to detect the novel images that deviate from previous patterns triggering particular vigilance on the part of the human agent. Due to the importance of this function, allowing computers to detect anomalies is a key task for artificial intelligence.

As a motivational example, let us consider the setting of an assembly-line fault detection. Assembly lines manufacture many instances of a particular product. Most products are normal and fault-free. Unfortunately, on isolated occasions, the manufactured products contain some faults e.g. dents, wrong labels or part duplication. As reputable manufacturers strive to keep a consistent quality of products, prompt detection of the faulty products is very valuable. As mentioned earlier, humans are quite adept at anomaly detection, however having a human operator oversee every product manufactured by the assembly line has several key limitations: i) high wages earned by skilled human operators ii) limited human attention span (\cite{green1999appropriate} states this can be as low as $20$ minutes!) iii) a human operator cannot be replicated between different assembly lines. iv) different operators typically do not maintain a consistent quality level. Anomaly detection therefore calls for computer vision solutions.

Although visual anomaly detection is very valuable, it is also quite challenging. One challenge common to all anomaly detection methods is the unexpectedness of anomalies. Typically in supervised classification, test classes come from the a similar distribution to the train data. In most anomaly detection settings, the distribution of anomalies is not observed during training time. Different anomaly detection methods differ by the way the anomalies are observed at training time. In this paper, we deal with the setting where at training time only normal data (but no anomalies) are observed. This is a practically useful setting, as obtaining normal data (e.g. products that contain no faults) is usually easy. This setting is sometimes called semi-supervised (\cite{chandola2009anomaly}). As this notation is ambiguous, we shall refer to this setting as the normal-only training setting. An easier scenario is fully supervised i.e. both normal and anomalous examples are presented with labels during training. As this training setting is  similar to standard supervised classification, a mature task with effective solutions, it will not be dealt with in this work.  

Another challenge particular to visual anomaly detection (rather than non-image anomaly detection methods) is the localization of anomalies i.e. segmenting the parts of the image which the algorithm deems anomalous. This is very important for the explainability of the decision made by the algorithm as well as for building trust between operators and novel AI systems. It is particularly important for anomaly detection, as the objective is to detect novel changes not seen before which humans might not be familiar with. In this case, the computer may teach the human operator of the existence of new anomalies or alternatively the human may decide that an anomaly is not of interest thus not rejecting the product, resulting in cost-saving

We present a new method for solving the task of sub-image anomaly detection and segmentation. Our method does not require an extended training stage, it is fast, robust and achieves state of the art performance. Out methods consists of several stages: i) image feature extraction using a pre-trained deep neural network (e.g. an ImageNet trained ResNet) ii) nearest neighbor retrieval of the nearest K normal images to the target iii) finding dense pixel-level correspondence between the target and the normal images, target image regions that do not have near matches in the retrieved normal images are labeled as anomalous. Our method is extensively evaluated on an industrial product dataset (MVTech) as well as a surveillance dataset in a campus setting (Shanghai Tech Campus). Our method achieves state-of-the-art performance both on image-level and pixel-level anomaly detection.   

\section{Previous Work}
\label{sec:prev}

Anomaly detection has attracted a large body of work over the last several decades. We present an overview of image-level and sub-image anomaly detection methods.

\textit{Image-level methods:} We review methods that detect if an image is anomalous that are not particularly designed for segmenting the anomaly within the image. Note that many of these methods are not specific to images. There are three main classes of methods for image-level anomaly detection: reconstruction-based, distribution-based and classification-based. 

Reconstruction-based methods learn a set of basis functions on the training data, and attempt to reconstruct the test image using a sparse set of these basis functions. If the test image cannot be faithfully reconstructed using the basis functions, it is denoted as anomalous, as it is likely that it came from a different basis from that of the normal training data. Different methods vary in terms of the set of basis functions and loss functions they use. Popular choices of basis functions include: K-means \cite{hartigan1979algorithm}, K nearest neighbors (kNN) \cite{eskin2002geometric}, principal component analysis (PCA) \cite{jolliffe2011principal}. The loss functions used vary between simple vector metrics such as Euclidean or $L_1$ losses and can use more complex perceptual losses such as structural similarity (SSIM) \cite{wang2004image}. Recently deep learning methods have broadened the toolbox of reconstruction-based methods. Principal components have been extended to non-linear functions learned by autoencoders \cite{sakurada2014anomaly}, including both denoising as well as variational autoencoders (VAEs). Deep perceptual loss functions \cite{zhang2016colorful} significantly improve over traditional perceptual loss functions. The main disadvantages of reconstruction-based loss functions are: i) sensitivity to the particular loss measure used for evaluating the quality of reconstruction, making their design non-obvious and hurting performance ii) determining the correct functional basis. 

The second class of methods is distribution-based. The main principle is to model the probability density function (PDF) of the distribution of the normal data. Test samples are evaluated using the PDF, and test samples with low probability density values are designated as anomalous. Different distribution-based methods differ by the distributional assumptions that they make, the approximations used to estimate the true PDF, and by the training procedure. Parametric methods include Gaussian or mixture of Gaussians (GMM). Kernel density estimation \cite{latecki2007outlier} is a notable non-parametric method. Nearest neighbors  \cite{eskin2002geometric} can also be seen as a distributional (as it performs density estimation), but note that we also designated it a reconstruction-based method. Recently deep learning methods have improved performance, particularly by mapping high-dimensional data distributions into a lower and denser space. PDF estimation is typically easier in lower dimensional spaces. Learning the deep projection and distributional modeling can be done jointly as done by \cite{zong2018deep}. Another recent development, adversarial training, was also applied to anomaly detection e.g. ADGAN \cite{deecke2018anomaly}. Although in principle distributional-methods are very promising, they suffer from some critical drawbacks: i) real image data rarely follows simple parametric distributional assumptions ii) non-parametric methods have high sample complexity and often require large training set that is often not available in practice.

Recently, classification-based methods have achieved dominance for image-level anomaly detection. One such paradigm is one-class support vector machines (OC-SVM) \cite{scholkopf2000support}. One of its most successful variants is support vector data description (SVDD) \cite{tax2004support} which can be seen as a finding the minimal sphere which contains at least a given fraction of the data. These methods are very sensitive to the feature space used giving rise to both kernel methods as well as deep methods \cite{ruff2018deep} for learning features. Another set of methods is based on self-supervised learning. Golan and El-Yaniv \cite{golan2018deep} proposed a RotNet-based \cite{gidaris2018unsupervised} approach, which performs geometric transformations on the input data and trains a network that attempts to recognize the transformation used. They use the idea that the trained classifier will generalize well to new normal images but not to anomalous images allowing it to be used as an anomaly detection criterion. Hendrycks et al. \cite{hendrycks2018deep} improved the architecture and training procedure achieving strong performance. Bergman and Hoshen \cite{bergman2020classification} combined this work with an SVDD type criterion and extended it to non-image data. Very recently Bergman et al. \cite{bergman2020deep} showed that the features learned using such self-supervised methods are not competitive with generic ImageNet-based feature extractors. A simple method based on kNN (or efficient approximations) significantly outperformed such self-supervised methods. 

\textit{Sub-image methods:} The methods previously described tackled the task of classifying a whole image as normal or anomalous, and most of the techniques were not specific to images. The task of segmenting the particular pixels containing anomalies is special to images and has achieved far less attention from the deep learning community. Napoletano et al. \cite{napoletano2018anomaly} extracted deep features from small overlapping patches, and used a K-means based classifier over dimensionality reduced features. Bergmann et al. \cite{bergmann2019mvtec} evaluated both a ADGAN and autoencoder approaches on MVTech finding complementary strengths. More recently, Venkataramanan et al. \cite{venkataramanan2019attention} used an attention-guided VAE approach combining multiple methods (GAN loss \cite{goodfellow2014generative}, GRADCAM \cite{selvaraju2017grad}). Bergmann et al. \cite{bergmann2019uninformed} used a student-teacher based autoencoder approach employing pre-trained ImageNet deep features (still requiring an expensive training stage). In this work, we present a novel sub-image alignment approach which is more accurate, faster, more stable than previous methods and does not require a dedicated training stage. To support research on sub-image anomaly detection, high quality datasets for evaluating this task have been introduced, such as: MVTech \cite{bergmann2019mvtec} - a dataset simulating an industrial fault detection where the objective is to detect parts of images of products that contain faults such as dents or missing parts. The ShanghaiTech Campus dataset \cite{luo2017revisit} - a dataset simulating a surveillance setting where cameras observe a busy campus and the objective is to detect anomalous objects and activities such as fights. Hendrycks et al. \cite{hendrycks2019benchmark} also proposed a new dataset containing anomalies such as road hazards. We evaluate our work on the two most used datasets, MVTech and ShanghaiTech Campus (STC).     

\section{Correspondence-based Sub-Image Anomaly Detection}
\label{sec:method}

We present our method for sub-image anomaly detection. Our method consists of several parts: i) image feature extraction ii) K nearest neighbor normal image retrieval iii) pixel alignment with deep feature pyramid correspondences.

\subsection{Feature Extraction} 
\label{sec:method:extract} 

The first stage of our method is the extraction of strong image level features. The same features are later used for pixel-level image alignment. There are multiple options for extracting features. The most commonly used option is self-supervised feature learning, that is, learning features from scratch directly on the input normal images. Although it is an attractive option, it is not obvious that the features learned on small training datasets will indeed be sufficient for serving as high-quality similarity measures. The analysis performed in Bergman et al. \cite{bergman2020classification} illustrates that self-supervised features underperform ImageNet-trained ResNet features for the purposes of anomaly detection. We therefore used a ResNet feature extractor pre-trained on the ImageNet dataset. As image-level features we used the feature vector obtained after global-pooling the last convolutional layer. Let us denote the global feature extractor $F$, for a given image $x_i$, we denote the extracted features $f_i$:

\begin{equation}
    \label{eq:extract}
    f_i = F(x_i)
\end{equation}

At initialization, the features for all training images (which are all normal) are computed and stored. At inference, only the features of the target image are extracted. 

\subsection{K Nearest Neighbor Normal Image Retrieval} 
\label{sec:method:retrieval} 

The first stage in our method is determining which images contain anomalies using DN$2$ \cite{bergman2020classification}. For a given test image $y$, we retrieve its $K$ nearest normal images from the training set, $N_k(f_y)$. The distance is measured using the Euclidean metric between the image-level feature representations.

\begin{equation}
    \label{eq:knn1}
    d(y) = \frac{1}{K} \sum_{f \in N_K(f_y)}{\|f - f_y\|^2}
\end{equation}

Images are labelled at this stage as normal or anomalous. Positive classification is determined by verifying if the kNN distance is larger than a threshold $\tau$. It is expected that most images are normal, and only few images are designated as anomalous.

\subsection{Sub-image Anomaly Detection via Image Alignment} 
\label{sec:method:alignment} 

After being labelled as anomalous at the image-level stage, the objective is to locate and segment the pixels of one or multiple anomalies. In the case that the image was falsely classified as anomalous, our objective would be to mark no pixels as anomalous.

As a motivational idea, let us consider aligning the test image to a retrieved normal image. By finding the differences between the test and normal image, we would hope to detect the anomalous pixels. This naive method has several flaws i) assume that there are multiple normal parts the object may possibly consist of, alignment to particular normal images may fail ii) for small datasets or objects that experience complex variation, we may never in fact find a normal training image which is similar to the test image in every respect triggering false positive detections  iii) computing the image difference would be very sensitive to the loss function being used.

To overcome the above issues, we present a multi-image correspondence method. We extract deep features at every pixel location $p \in P$ using feature extractor $F(x_i, p)$ of the relevant test and normal training images. The details of the feature extractor will be described in Sec.~\ref{sec:method:pyramid}. We construct a gallery of features at all pixel locations of the $K$ nearest neighbors $G = \{F(x_1, p)|p \in P\} \cup \{F(x_2, p)|p \in P\} \}..\cup \{F(x_K, p)|p \in P\} \}$. The anomaly score at pixel $p$, is given by the average distance between the features $F(y,p)$ and its $\kappa$ nearest features from the gallery $G$. The anomaly score of pixel $p$ in target image $y$ is therefore given by:

\begin{equation}
    \label{eq:knn2}
    d(y, p) = \frac{1}{\kappa} \sum_{{f \in  N_{\kappa}}(F(y, p))}{\|f - F(y, p)\|^2}
\end{equation}

For a given threshold $\theta$, a pixel is determined as anomalous if $d(y, p) > \theta$, that is, if we cannot find a closely corresponding pixel in the $K$ nearest neighbor normal images.  

\subsection{Feature Pyramid Matching} 
\label{sec:method:pyramid} 

Alignment by dense correspondences is an effective way of determining the parts of the image that are normal vs. those that are anomalous. In order to perform the alignment effectively, it is necessary to determine the features for matching. As in the previous stage, our method uses features from a pre-trained deep ResNet CNN. The ResNet results in a pyramid of features. Similarly to image pyramids, earlier layers (levels) result in higher resolution features encoding less context. Later layers encode lower resolution features which encode more context but at lower spatial resolution. To perform effective alignment, we describe each location using features from the different levels of the feature pyramid. Specifically, we concatenate features from the output of the last $M$ blocks, the results for different numbers of $M$ is shown in the experimental section. Our features encode both fine-grained local features and global context. This allows us to find correspondences between the target image and $K \ge 1$ normal images, rather than having to explicitly align the images, which is more technically challenging and brittle. Our method is scalable and easy to deploy in practice. We will show in Sec.~\ref{sec:exp} that our method achieves the state-of-the-art sub-image anomaly segmentation accuracy.

\begin{figure}[t]
\begin{center}
    \begin{tabular}{cccc}
     \includegraphics[scale=0.33]{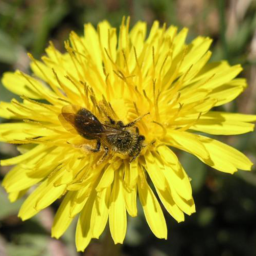} &
   \includegraphics[scale=0.376]{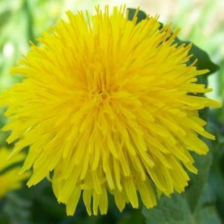} &
   \includegraphics[scale=0.33]{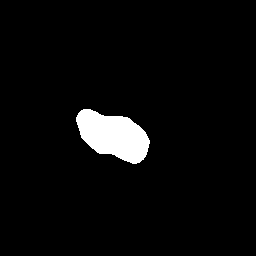} &
   \includegraphics[scale=0.33]{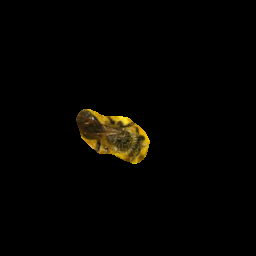} \\
        \includegraphics[scale=0.33]{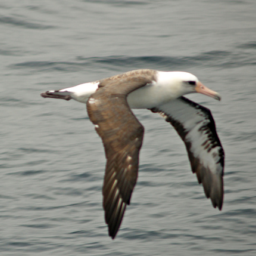} &
   \includegraphics[scale=0.376]{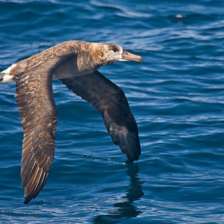} &
   \includegraphics[scale=0.33]{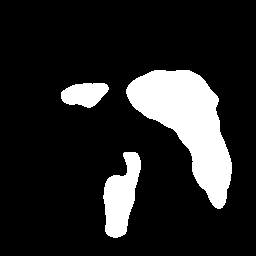} &
   \includegraphics[scale=0.33]{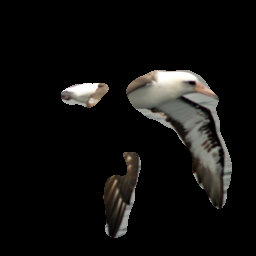} \\
       \includegraphics[scale=0.33]{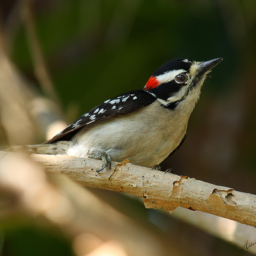} &
   \includegraphics[scale=0.376]{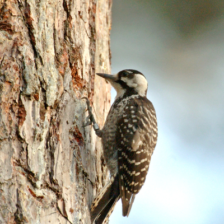} &
   \includegraphics[scale=0.33]{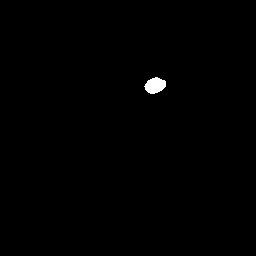} &
   \includegraphics[scale=0.33]{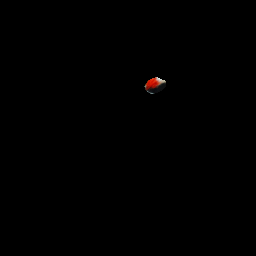}
   
    \end{tabular}
    \end{center}
    \caption{ An evaluation of SPADE on detecting anomalies between flowers with or without insects (taken from one category of 102 Category Flower Dataset \cite{nilsback2008automated}) and bird varieties (taken from Caltech-UCSD Birds 200) \cite{welinder2010caltech}.   (left to right) i) An anomalous image ii) The retrieved top normal neighbor image iii) The mask detected by SPADE iv) The predicted anomalous image pixels. SPADE was able to detect the insect on the anomalous flower (top), the white colors of the anomalous albatross (center) and the red spot on the anomalous bird (bottom). }
    \label{fig:flower}
\end{figure}

\subsection{Implementation Details} 
\label{sec:method:imp} 

In all experiments, we use a Wide-ResNet$50\times2$ feature extractor, which was pre-trained on ImageNet. MVTec images were resized to $256 \times 256$ and cropped to $224 \times 224$. STC images were resized to $256 \times 256$. Due to the large size of STC dataset, we subsampled its training data to roughly $5000$ images. To be comparable with \cite{venkataramanan2019attention}, we subsampled the STC test set by a factor of $5$. All metrics were calculated at $256 \times 256$ image resolution, and we used cv2.INTERAREA for resizing when needed. Unless otherwise specified, we used features from the ResNet at the end of the first block ($56 \times 56$), second block ($28 \times 28$) and third block ($14 \times 14$), all with equal weights. We used $K = 50$ nearest neighbours for the MVTtec experiments and $K = 1$ nearest neighbours for the STC experiments (due to the runtime considerations). In all experiments we used $\kappa=1$.

After achieving the pixel-wise anomaly score for each image, we used smoothed the results with a Gaussian filter ($\sigma = 4$).

\section{Experiments}
\label{sec:exp}

We perform an extensive evaluation of our method against the state-of-the-art in sub-image anomaly detection.

\subsection{MVTec}

To simulate anomaly detection in industrial settings, \cite{bergmann2019mvtec} Bregmann et al. presented a dataset consisting of images from $15$ different classes. $5$ classes consist of textures such as wood or leather. The other $10$ classes contain objects (mostly rigid). For each class, the training set is composed of normal images. The test set is composed of normal images as well as images containing different types of anomalies. This dataset therefore follows the standard protocol where no anomalous images are used in training. The anomalies in this dataset are more finegrained than those typically used in the literature e.g. in CIFAR10 evaluation, where anomalous images come from a completely different image category. Instead, anomalies take the form of a slightly scratched hazelnut or a lightly bent cable. As the anomalies are at the sub-image level, i.e. only affect a part of the image, the dataset provides segmentation maps indicating the precise pixel positions of the anomalous regions. 

\begin{figure}[t]
\begin{center}
    \begin{tabular}{cccc}
   \includegraphics[scale=0.33]{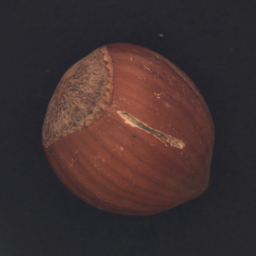} &
   \includegraphics[scale=0.376]{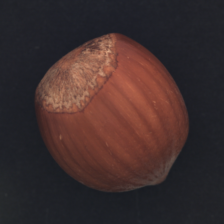} &
   \includegraphics[scale=0.33]{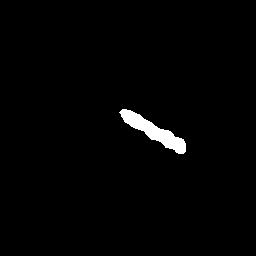} &
   \includegraphics[scale=0.33]{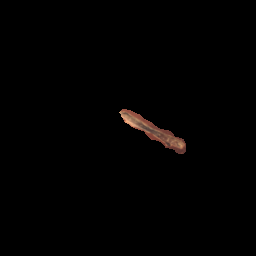} \\
        \includegraphics[scale=0.33]{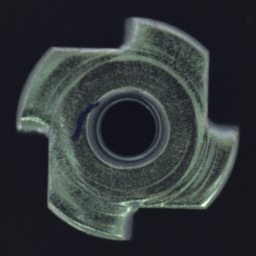} &
   \includegraphics[scale=0.376]{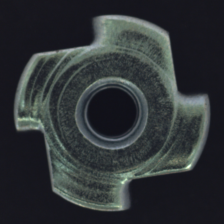} &
   \includegraphics[scale=0.33]{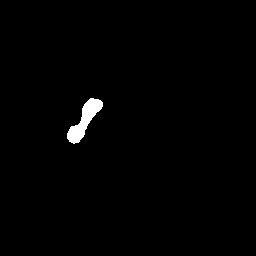} &
   \includegraphics[scale=0.33]{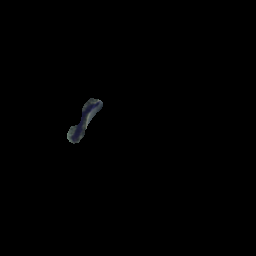}

    \end{tabular}
    \end{center}
    \caption{(left to right) i) An anomalous image ii) The retrieved top normal neighbor image iii) The mask detected by SPADE iv) The predicted anomalous image pixels. We can see how in this example, SPADE detects the anomalous image region by finding the correspondence with the nearest-neighbor image. The anomalous parts did not have correspondences in the normal image and were therefore detected. }
    \label{fig:hazelnut}
\end{figure}

An example of the operation of our method on the MVTec dataset can be observed in Fig.~\ref{fig:hazelnut}. The anomalous object (a hazelnut) contain a scratch. The retrieved nearest neighbor normal image, contains a complete nut without scratches. By search for correspondences between the two images, our method is able to find correspondences for the normal image regions but not for the anomalous region. This results in an accurate detection of the anomalous region of the image. The anomalous images pixels are presented on the right-most image.  

We compared our method against several methods that were introduced over the last several months, as well as longer standing baseline such as OCSVM and nearest neighbors. For each setting, we compared against the methods that reported the suitable metric.

We first compare the quality of deep nearest neighbor matching as a means for finding anomalous images. This is computed by the distance between the test image and the nearest neighbor normal images. Larger distances indicate more anomalous images. We compared the ROC area under the curve (ROCAUC) of the first step of our method and other state-of-the-art methods for image level anomaly detection. We report the average ROCAUC across the $15$ classes. Please note that the first stage of our method is identical with DN2 \cite{bergman2020deep}. This comparison is important as it verifies if deep nearest neighbors are effective on these dataset. The comparison is presented in Tab.~\ref{tab:mvtec_image_table}. Our method is shown to outperform a range of state-of-the-art methods utilizing a range of self-supervised anomaly detection learning techniques. This gives evidence that deep features trained on ImageNet (which is very different from ImageNet dataset) are very effective even on datasets that are quite different from ImageNet.

\begin{table}[t]
  \centering
  \caption{Image-level anomaly detection accuracy on MVTec (Average ROCAUC $\%$)}
  \label{tab:mvtec_image_table}

    \begin{tabular}{lccccc}
    \toprule      

   & Geom \cite{golan2018deep} & GANomaly \cite{akcay2018ganomaly} & $AE_{L2}$ & ITAE \cite{huang2019inverse} & SPADE\\
    \midrule
   Average & 67.2 & 76.2 & 75.4 & 83.9 & \textbf{85.5} \\
	 \bottomrule
    \end{tabular}
\end{table}

We proceed to evaluate our method on the task of pixel-level anomaly detection. The objective here is to segment the particular pixels that contain anomalies. We evaluate our method using two established metrics. The first is per-pixel ROCAUC. This metric is calculated by scoring each pixel by the distance to its $K$ nearest correspondences. By scanning over the range of thresholds, we can compute the pixel-level ROCAUC curve. The anomalous category is designated as positive. It was noted by several previous works that ROCAUC is biased in favor of large anomalies. In order to reduce this bias, Bergmann et al \cite{bergmann2019uninformed} propose the PRO (per-region overlap) curve metric. They first separate anomaly masks into their connected components, therefore dividing them into individual anomaly regions. By changing the detection threshold, they scan over false positive rates (FPR), for each FPR they compute PRO i.e. the proportion of the pixels of each region that are detected as anomalous. The PRO score at this FPR is the average coverage across all regions. The PRO curve metric computes the integral across FPR rates from $0$ to $0.3$. The PRO score is the normalized value of this integral. 

\begin{table}[t]
  \centering
  \caption{Sub-Image anomaly detection accuracy on MVTec (ROCAUC $\%$)}
  \label{tab:mvtec_pixel_roc}

    \begin{tabular}{lcccccccc}
    \toprule      

   & $AE_{SSIM}$  &  $AE_{L2}$  & AnoGAN  & CNN Dict & TI & VM & CAVGA-$R_u$  & SPADE\\
    \midrule
Carpet &	87	&	59	&	54	&	72	&	88	&	-	& -	& 97.5	\\
Grid &	94	&	90	&	58	&	59	&	72	&	-	&-	& 93.7	\\
Leather &	78	&	75	&	64	&	87	&	97	&	-	&-	& 97.6	\\
Tile &	59	&	51	&	50	&	93	&	41	&	-	&	-& 87.4	\\
Wood &	73	&	73	&	62	&	91	&	78	&	-	&-	& 88.5	\\
Bottle &	93	&	86	&	86	&	78	&	-	&	82	& -	& 98.4	\\
Cable &	82	&	86	&	78	&	79	&	-	&	-	& -	& 97.2	\\
Capsule &	94	&	88	&	84	&	84	&	-	&	76	& -	& 99.0	\\
Hazelnut &	97	&	95	&	87	&	72	&	-	&	-	& -	& 99.1	\\
Metal nut &	89	&	86	&	76	&	82	&	-	&	60	&	-& 98.1	\\
Pill &	91	&	85	&	87	&	68	&	-	&	83	&-	& 96.5	\\
Screw &	96	&	96	&	80	&	87	&	-	&	94	&	-& 98.9	\\
Toothbrush &	92	&	93	&	90	&	77	&		&	68	&-	& 97.9	\\
Transistor &	90	&	86	&	80	&	66	&	-	&	-	&-	& 94.1	\\
Zipper &	88	&	77	&	78	&	76	&	-	&	-	&-	& 96.5	\\
\midrule														
Average & 87	&	82	&	74	&	78	&	75	&	77	& 89	& \textbf{96.0}	\\
	 \bottomrule
    \end{tabular}
\end{table}

In Tab.~\ref{tab:mvtec_pixel_roc}, we compare our methods on the per-pixel ROCAUC metric against state-of-the-art results reported by Bergmann et al. \cite{bergmann2019mvtec} as well as newer results by Venkataramanan et al. \cite{venkataramanan2019attention}. Most of the methods use different varieties of autoencoders, including the top-performer CAVGA-$R_u$. Our method significantly outperforms all methods. This attest to the strength of our pyramid based correspondence approach. 

\begin{table}[t]
  \centering
  \caption{Sub-Image anomaly detection accuracy on MVTec (PRO $\%$)}
  \label{tab:mvtec_pixel_pro}

    \begin{tabular}{lcccccccc}
    \toprule      

    & Student & 1-NN & OC-SVM  & $\ell_2$-AE & VAE & SSIM-AE  & CNN-Dict & SPADE \\
    \midrule
Carpet	&			69.5	&	51.2	&	35.5	&	45.6	&	50.1	&	64.7	&	46.9	&	94.7		\\
Grid	&			81.9	&	22.8	&	12.5	&	58.2	&	22.4	&	84.9	&	18.3	&	86.7		\\
Leather	&			81.9	&	44.6	&	30.6	&	81.9	&	63.5	&	56.1	&	64.1	&	97.2		\\
Tile	&			91.2	&	82.2	&	72.2	&	89.7	&	87.0	&	17.5	&	79.7	&	75.9		\\
Wood	&			72.5	&	50.2	&	33.6	&	72.7	&	62.8	&	60.5	&	62.1	&	87.4		\\
Bottle	&			91.8	&	89.8	&	85.0	&	91.0	&	89.7	&	83.4	&	74.2	&	95.5		\\
Cable	&			86.5	&	80.6	&	43.1	&	82.5	&	65.4	&	47.8	&	55.8	&	90.9		\\
Capsule	&			91.6	&	63.1	&	55.4	&	86.2	&	52.6	&	86.0	&	30.6	&	93.7		\\
Hazelnut	&			93.7	&	86.1	&	61.6	&	91.7	&	87.8	&	91.6	&	84.4	&	95.4		\\
Metal nut	&			89.5	&	70.5	&	31.9	&	83.0	&	57.6	&	60.3	&	35.8	&	94.4		\\
Pill	&			93.5	&	72.5	&	54.4	&	89.3	&	76.9	&	83.0	&	46.0	&	94.6		\\
Screw	&			92.8	&	60.4	&	64.4	&	75.4	&	55.9	&	88.7	&	27.7	&	96.0		\\
Toothbrush	&			86.3	&	67.5	&	53.8	&	82.2	&	69.3	&	78.4	&	15.1	&	93.5		\\
Transistor	&			70.1	&	68.0	&	49.6	&	72.8	&	62.6	&	72.5	&	62.8	&	87.4		\\
Zipper	&			93.3	&	51.2	&	35.5	&	83.9	&	54.9	&	66.5	&	70.3	&	92.6		\\
\midrule																				
Average	&			85.7	&	64	&	47.9	&	79	&	63.9	&	69.4	&	51.5	&	\textbf{91.7}		\\																
	 \bottomrule
    \end{tabular}
\end{table}

In Tab.~\ref{tab:mvtec_pixel_pro}, we compare our method in terms of PRO. As explained above, this is another per-pixel accuracy measure which gives larger weight to anomalies which cover few pixels. Our method is compared with the auto-encoder with pre-trained features based approach of Bregmann et al. \cite{bergmann2019uninformed} and the baselines presented in their paper. Our approach achieves significantly better results than all previous methods. We note than Bregmann et al also presented an ensemble approach with better results. While our method does not use ensembles (which will probably improve our method too), we outperform the ensemble approach as well. We present more qualitative results of our method in Fig.~\ref{fig:flower} that show that our method is able to recover accurate masks of the anomalous regions. 

\subsection{Shanghai Tech Campus Dataset}

\begin{table}[t]
  \centering
  \caption{Image-level anomaly detection accuracy on STC (Average ROCAUC $\%$)}
  \label{tab:mstc_image}

    \begin{tabular}{ccccccc}
    \toprule      

   TSC \cite{luo2017revisit}  & StackRNN \cite{luo2017revisit}  & AE-Conv3D \cite{zhao2017spatio}  & MemAE \cite{gong2019memorizing}  & AE(2D) \cite{hasan2016learning}  & ITAE \cite{huang2019inverse}  & SPADE\\
    \midrule
   67.9 & 68.0 & 69.7 & 71.2 & 60.9 & \textbf{72.5} & 71.9\\
	 \bottomrule
    \end{tabular}
\end{table}

We evaluate our method on the Shanghai Tech Campus dataset. It simulates a surveillance setting, where the input consists of videos captured by surveillance cameras observing a busy campus. The dataset contains $12$ scenes, each scene consists of training videos and a smaller number of test images. The training videos do not contain anomalies while the test videos contain normal and anomalous images. Anomalies are defined as pedestrians performing non-standard activities (e.g. fighting) as well as any moving object which is not a pedestrian (e.g. motorbikes). 

\begin{table}[t]
  \centering
  \caption{Pixel-level anomaly detection accuracy on STC (Average ROCAUC $\%$)}
  \label{tab:mstc_table}

    \begin{tabular}{cccc}
    \toprule      

   $AE_{L2}$ & $AE_{SSIM}$ & CAVGA-$R_u$ \cite{venkataramanan2019attention} & SPADE\\
    \midrule
   74 & 76 & 85 & \textbf{89.9} \\
	 \bottomrule
    \end{tabular}
\end{table}

We began by evaluating our first stage for detecting image-level anomalies against other state-of-the-art methods. We show in Tab.~\ref{tab:mstc_image} that our first stage has comparable performance to the top performing method \cite{huang2019inverse}. More interestingly, we compare in Tab.~\ref{tab:mstc_table} the pixel-level ROCAUC performance with the best reported method, CAVGA-$R_u$ \cite{venkataramanan2019attention}. Our method significantly outperforms the best reported method by a significant margin. Note that we compared to the best method that did not use anomaly supervision, as we do not use it and as anomaly supervision is often not available in practice.

%\begin{figure}

 %   \begin{tabular}{ccc}

  % \includegraphics[scale=0.457]{images/images_NN/NormalNN.png} &
  % \includegraphics[scale=0.4]{images/images_NN/Normal.png} \\
  %  \end{tabular}
  %  \caption{NN with anomaly and train}
  %  \label{fig:small_depth}
%\end{figure}

\begin{figure}
\begin{center}
    \begin{tabular}{ccc}
    \includegraphics[scale=0.25]{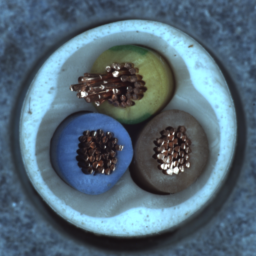} &
   \includegraphics[scale=0.25]{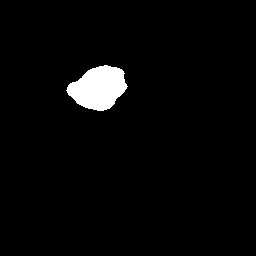} &
   \includegraphics[scale=0.25]{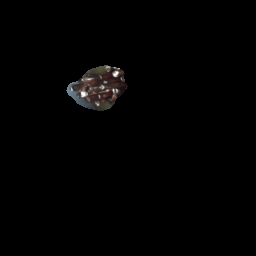} \\
   \includegraphics[scale=0.25]{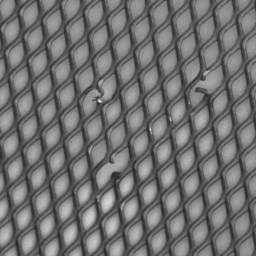} &
   \includegraphics[scale=0.25]{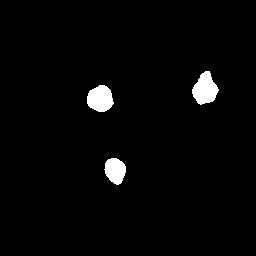} &
   \includegraphics[scale=0.25]{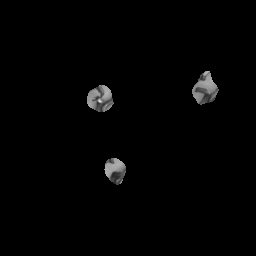} \\
   \includegraphics[scale=0.25]{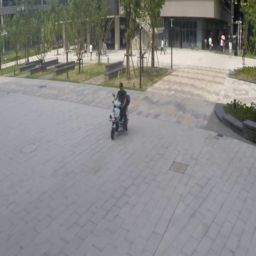} &
   \includegraphics[scale=0.25]{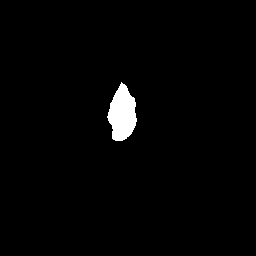} &
   \includegraphics[scale=0.25]{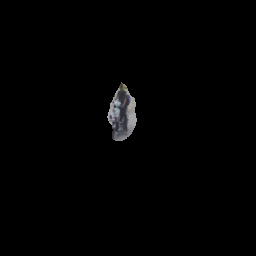} \\
    \end{tabular}
    \end{center}
    \caption{(top rows) Anomaly detection on images from the Cable and Grid categories of the MVTec dataset (bottom) Detecting bike anomaly on the STC dataset.}
    \label{fig:small_depth}
\end{figure}

\subsection{Ablation Study}

We conduct an ablation study on our method in order to understand the relative performance of its different parts. In Tab.~\ref{tab:ablation:level}, we compare using different level of the feature pyramid. We experienced that using activations of too high resolution ($56 \times 56$) significantly hurts performance (due to limited context) while using the higher levels on their own results in diminished performance (due to lower resolution). Using a combination of all features in the pyramid results in the best performance. In Tab.~\ref{tab:K_ablation}, we compared using the top $K$ neighboring normal images as performed by our first stage vs. choosing them randomly from the dataset. We observe that choosing the kNN images improves performance. This does not affect all classes equally. As an example, we report the numbers for the class "Grid" which has much variation between images. For this category, using the kNN images results in much better performance than randomly choosing K images.  

\begin{table}[t]
  \centering
  \caption{Pyramid level ablation for sub-image anomaly detection accuracy on MVTec   (PRO $\%$)  }
  \label{tab:ablation:level}

    \begin{tabular}{lcccc}
    \toprule      

   Used layers size: & (14) & (28) & (56)  & SPADE \\
    \midrule
Carpet	&	93.5	&	93.4	&	91.0	&	94.7	\\
Grid	&	80.9	&	88.0	&	89.1	&	86.7	\\
Leather	&	96.6	&	97.5	&	97.3	&	97.2	\\
Tile	&	74.5	&	65.9	&	73.8	&	75.9	\\
Wood	&	84.7	&	87.7	&	87.5	&	87.4	\\
Bottle	&	93.7	&	94.7	&	88.3	&	95.5	\\
Cable	&	89.3	&	87.3	&	73.5	&	90.9	\\
Capsule	&	90.5	&	92.8	&	91.4	&	93.7	\\
Hazelnut	&	92.7	&	95.8	&	96.2	&	95.4	\\
Metal nut	&	91.3	&	93.1	&	86.1	&	94.4	\\
Pill	&	89.2	&	94.4	&	96.3	&	94.6	\\
Screw	&	90.7	&	95.9	&	96.1	&	96.0	\\
Toothbrush	&	90.9	&	93.5	&	94.5	&	93.5	\\
Transistor	&	91.3	&	72.1	&	62.5	&	87.4	\\
Zipper	&	90.9	&	92.4	&	92.5	&	92.6	\\
\midrule									
Average	&	89.38	&	89.6	&	87.74	&	\textbf{91.7}	\\

	 \bottomrule
    \end{tabular}
\end{table}

\begin{table}[t]
  \centering
  \caption{Evaluating the effectiveness of our kNN retrieval state. We use here $10$ nearest neighbours, chosen according to stage $1$, or randomly selected. We also show the "Grid" class to indicate that stage $1$ is more important to some classes then others }
  \label{tab:K_ablation}

    \begin{tabular}{lcccc}
    \toprule      

   Stage 1: &  SPADE (10 Random)  & SPADE (10NN) \\
    \midrule
Grid		&	73.2	&	86.3	\\
\midrule											
Average 	&	89.2	&	\textbf{91.4}	\\

	 \bottomrule
    \end{tabular}
\end{table}

\section{Discussion}
\label{sec:disc}

\textbf{Anomaly detection via alignment:} Most current sub-image anomaly detection methods take the approach of learning a large parametric function for auto-encoding images, making the assumption that anomalous regions will not be reconstructed well. Although this approach does achieve some success, we take a much simpler approach. Similarly to image alignment methods and differently from other sub-image anomaly detection methods, our method does not require feature training and can work on very small datasets. A difference between our method and standard image alignment is that we find correspondences between the target image and parts of $K$ normal images, as opposed to an entire single normal image in simple alignment approaches. The connection with alignment methods, can help in speeding up our method e.g. by combining it with the PatchMatch \cite{barnes2009patchmatch} method which used locality for significant speedup of the kNN search.    

\textbf{The role of context for anomaly detection:} The quality of the alignment between the anomalous image and retrieved normal images is strongly affected by the quality of extracted features. Similarly to other works dealing with detection and segmentation, the context is very important. Local context is needed for achieving segmentation maps with high-pixel resolutions. Such features are generally found in the shallow layers of a deep neural networks. Local context is typically insufficient for alignment without understanding the global context i.e. location of the part within the object. Global context is generally found in the deepest layers of a neural network, however global context features are of low resolution. The combination of features from different levels allows both global context and local resolution giving high quality correspondences. The idea is quite similar to that in Feature Pyramid Networks \cite{lin2017feature}. 

\textbf{Optimizing runtime performance:} Our method is significantly reliant on the K nearest neighbors algorithm. The complexity of kNN scales linearly with the size of the dataset used for search which can be an issue when the dataset is very large or of high dimensionality. Our approach is designed to mitigate the complexity issues. First, we compute the initial image-level anomaly classification on global-pooled features which are $2048$ dimensional vectors. Such kNN computation can be achieved very quickly for moderate sized datasets and different speedup techniques (e.g. KDTrees) can be used for large scale datasets. The anomaly segmentation stage requires pixel-level kNN computation which is significantly slower than image-level kNN. However, our method limits the sub-image kNN search to only the K nearest neighbors of the anomalous image significantly limiting computation time. We assume that the vast majority of images are normal, therefore only a small fraction of images require the next stage of anomaly segmentation. Additionally, the anomaly segmentation stage is required for explainability and trust building with the human operators, but in many cases it is not time-critical therefore putting a laxer requirement on computation time. Our method is therefore quite suitable for practical deployment from a complexity and runtime perspective.

\textbf{Pre-trained vs. learned features:} Previous sub-image anomaly detection methods have either used self-learned features or a combination of self-learned and pre-trained images features. Self-learned approaches in this context, typically train an autoencoder and use its reconstruction error for anomaly detection. Other approaches have used a combination of pre-trained and self-learned methods e.g. methods that use perceptual losses and \cite{bergmann2019uninformed} which uses a pre-trained encoder. Our numerical results have shown that our method significantly outperforms such approaches. We believe that given the limited supervision and small dataset size in normal-only training set as tackled in this work, it is rather hard to beat very deep pre-trained networks. We therefore use pre-trained features and do not attempt to modify them. The strong results achieved by our method attest to the effectiveness of this approach. We believe that future work should focus on methods for finetuning the deep pre-trained features for this particular task and expect it it improve over our method. That not-withstanding the ease of deployment and generality of our approach should make it a good choice in many practical settings.

\section{Conclusion}
\label{sec:conc}

We presented a novel alignment-based method for detecting and segmenting anomalies inside images. Our method relies on K nearest neighbors of pixel-level feature pyramids extracted by pre-trained deep features. Our method consists of two stages, which are designed to achieve high accuracy and reasonable computational complexity. Our method was shown to outperform the strongest current methods on two realistic sub-image anomaly detection datasets, while being much simpler. The ease of deployment enjoyed by our method should make it a good candidate for practitioners. 

\section{Acknowledgements}
 This work was partly supported by the Federmann Cyber Security Research Center in conjunction with the Israel National Cyber Directorate.

\clearpage
\bibliographystyle{splncs04}
\bibliography{egbib}
\end{document}